\documentclass{article}

\usepackage[square,numbers,compress]{natbib}
\usepackage[usenames, dvipsnames]{color}

\usepackage[preprint]{nips_2018}

\usepackage[utf8]{inputenc} 
\usepackage[T1]{fontenc}    
\usepackage{hyperref}       
\usepackage{url}            
\usepackage{booktabs}       
\usepackage{amsfonts}       
\usepackage{nicefrac}       
\usepackage{microtype}      
\usepackage[pdftex]{graphicx} 
\usepackage{amsmath}
\usepackage{booktabs}
\usepackage{caption}
\usepackage{subcaption}
\usepackage[export]{adjustbox}
\usepackage{newfloat}
\usepackage{placeins}
\usepackage{algorithm}
\usepackage[noend]{algpseudocode}
\usepackage[symbol]{footmisc}
\usepackage{footnote}
\makesavenoteenv{tabular}
\makesavenoteenv{table}

\DeclareFloatingEnvironment[
  fileext = lofa,
  placement = htbp,
  name = Figure,
  within = section
  ]{figurea}
\usepackage{caption}
\usepackage[title]{appendix}

\title{Intermediate Level Adversarial Attack for Enhanced Transferability}

\makeatletter
\newcommand{\printfnsymbol}[1]{%
  \textsuperscript{\@fnsymbol{#1}}%
}
\makeatother

\author{
  Qian Huang\thanks{Equal contribution.} \\
  Cornell University \\
  \texttt{qh53@cornell.edu} \\
  \And
  Zeqi Gu\printfnsymbol{1} \\
  Cornell University \\
  \texttt{zg45@cornell.edu} \\
  \And
  Isay Katsman\thanks{Equal contribution.} \\
  Cornell University \\
  \texttt{isk22@cornell.edu} \\
  \AND
  Horace He\printfnsymbol{2} \\
  Cornell University \\
  \texttt{hh498@cornell.edu} \\
  \And
  Pian Pawakapan \\
  Cornell University \\
  \texttt{pp456@cornell.edu} \\
  \And
  Zhiqiu Lin \\
  Cornell University \\
  \texttt{zl279@cornell.edu} \\
  \And
  Serge Belongie \\
  Cornell University \\
  \texttt{sjb344@cornell.edu} \\
  \And
  Ser-Nam Lim \\
  Facebook AI \\
  \texttt{sernam@gmail.com} \\
}

\begin{document}

\maketitle
\renewcommand*{\thefootnote}{\arabic{footnote}}
\begin{abstract}
   Neural networks are vulnerable to adversarial examples, malicious inputs crafted to fool trained models. Adversarial examples often exhibit black-box transfer, meaning that adversarial examples for one model can fool another model. However, adversarial examples may be overfit to exploit the particular architecture and feature representation of a source model, resulting in sub-optimal black-box transfer attacks to other target models. This leads us to introduce the \textit{Intermediate Level Attack} (ILA), which attempts to fine-tune an existing adversarial example for greater black-box transferability by increasing its perturbation on a pre-specified layer of the source model. We show that our method can effectively achieve this goal and that we can decide a nearly-optimal layer of the source model to perturb without any knowledge of the target models.
\end{abstract}

\section{Introduction}

Adversarial examples for neural networks are small perturbations of inputs carefully crafted to fool trained models \cite{Szegedy2013IntriguingPO, Goodfellow2014ExplainingAH}. In the context of Convolutional Neural Networks (CNNs) \cite{Krizhevsky2012ImageNetCW}, adversarial examples are non-perceptible perturbations of natural images, and they have been well studied in terms of attack \cite{Goodfellow2014ExplainingAH, Carlini2017TowardsET, Papernot2017PracticalBA, Dong2017BoostingAA, Athalye2018ObfuscatedGG}. Concerns have been raised over the vulnerability of CNNs to these perturbations in real-world contexts where they are used for online content filtration systems and self-driving cars \cite{Eykholt2017RobustPA, Kurakin2016AdversarialEI}.

Moreover, our current understanding of these perturbations is quite limited. In particular, they have been known to exhibit black-box transfer, meaning that perturbations crafted to fool one model will also fool another model. Though some works have attempted to shed light on this phenomenon \cite{Tramr2017TheSO}, many questions remain unanswered. In our work, we aim to provide more insights on this subject by exploring how perturbations on a source model can be enhanced for greater black-box transfer. We attempt to do this by maximizing their effect on one of the source model's intermediate layers.

Our contributions are as follows:
\begin{itemize}
  \setlength\itemsep{1em}
  \item We propose a novel method, \textit{Intermediate Level Attack} (ILA), that enhances black-box adversarial transferability by increasing the perturbation on a pre-specified layer of a model.
  \item When attacking a model, it is generally effective to focus on perturbing the last layer. However, we show that when optimizing for black-box transfer it is better to emphasize perturbing an intermediate layer.
  \item Additionally, we provide a method for selecting an optimal layer for transferability using the source model alone, thus obviating the need for evaluation on transfer models during hyperparameter optimization.
\end{itemize}

\section{Motivation and Approach}
For the duration of this paper, we will focus on non-targeted image-dependent attacks. In a typical attack setting, most methods, such as I-FGSM \cite{Kurakin2016AdversarialEI}, PGD \cite{Madry2017TowardsDL}, and DeepFool \cite{MoosaviDezfooli2016DeepFoolAS}, primarily focus on attacking the last layer, with most loss objectives formulated directly in terms of cross-entropy with the softmax output and some desired discrete distribution. Although these attack methods produce transferable adversarial examples \cite{Tramr2017TheSO}, they are not inherently emphasizing this property. Our attack, on the other hand, focuses on enhancing transferability of the adversarial examples by perturbing intermediate layers.

Motivated by \cite{Zeiler2014VisualizingAU}, we view a CNN's convolutional layer as having learned a feature hierarchy, with the earliest layers having learned primitive features and latest layers having learned the most high-level features. We hypothesize that for a given model the low-level feature representations are similar to those of other models until they begin to diverge at a specific layer (when models learn different representations of high-level features). Thus, we aim to find and attack the latest layer at which the feature representations learned are still general enough to be found in other models. Adversarial examples crafted to attack such an intermediate layer will be more transferable as they effectively focus on attacking the internal representation that is common to most models.

Based on the above motivation, we propose the following attack, which we call \textit{Intermediate Level Attack} (ILA). We define $F_l(X)$ as the output of layer $l$ of a network $F$ given an input $X$. Given an adversarial example $X'$ generated by attack method $A$ for natural image $X$ (generated after $n$ iterations)\footnote{The attack does not have to be iterative, but in this paper we focus on iterative attacks.}, and a specific layer $l$ of a network $F$, we aim to produce an $X''$ that solves:

\begin{equation} \label{eq:main}
\begin{gathered}
\max_{||X'' - X||_\infty < \epsilon} \overbrace{\alpha \cdot \frac{||\Delta Y''_l||_2}{||\Delta Y'_l||_2}}^{\text{maximize disturbance}} + \overbrace{\frac{\Delta Y''_l}{||\Delta Y''_l||_2} \bullet \frac{\Delta Y'_l}{||\Delta Y'_l||_2}}^{\text{maintain original direction}} \\ \\
\text{where $\Delta Y'_l = F_l(X') - F_l(X)$} \\
\text{and $\Delta Y''_l = F_l(X'') - F_l(X)$}
\end{gathered}
\end{equation}

In practice, we iterate $m$ times to attain this objective (full algorithm given in Appendix A). Note that the layer $l$ and the loss weight $\alpha$ are hyperparameters of this attack. Also, note that $\bullet$ represents a dot product. The above attack assumes that $X'$ is a pre-generated adversarial example. As such, the attack can be viewed as a fine-tuning of the adversarial example $X'$. We fine-tune for greater norm of the output difference at layer $l$ (which we hope will be conducive to greater transferability) while attempting to preserve the output difference's direction to avoid destroying the original adversarial structure.

\section{Results}

We start by showing that our method increases transferability for various base attack methods, including I-FGSM \cite{Kurakin2016AdversarialEI} and I-FGSM with momentum \cite{Dong2017BoostingAA}. We test on a variety of models, namely: ResNet18 \cite{He2016DeepRL}, SENet18 \cite{Hu2017SqueezeandExcitationN}, DenseNet121\cite{Huang2017DenselyCC} and GoogLeNet \cite{Szegedy2015GoingDW}. Architecture details are specified in Appendix B; note that in the below results sections, instead of referring to the architecture specific layer names, we refer to layer indices (e.g. $l=0$ is the last layer of the first block). Our models are trained on CIFAR-10 \cite{Krizhevsky2009LearningML} with the code and hyperparameters in \cite{Liu2018} to final test accuracies of $94.8\%$ for ResNet18, $94.6\%$ for SENet18, $95.6\%$ for DenseNet121, and $94.9\%$ for GoogLeNet.

For a fair comparison, we use the output of an attack $A$ run for 20 iterations as a baseline. ILA is run for $10$ iterations starting from the output of attack $A$ after $10$ iterations. The learning rate is set to $0.002$ for both I-FGSM and I-FGSM with momentum\footnote{Tuning the learning rate does not substantially affect transferability, as shown in Appendix E.}. The learning rate for ILA is set to $1.0$ (this value is tuned to exhibit near optimal averaged attack strength on transferred models). Finally, we limit ourselves to generating adversarial examples $X''$ for natural images\footnote{Our CIFAR-10 images are normalized to be in the range $[-1,1]$ via the transform $\frac{x / 255 - 0.5}{0.5}$. Results for different values of $L_\infty$ are given in Appendix D.} $X$ such that $||X''-X||_\infty < \epsilon$ (we choose $\epsilon = 0.03$).

To evaluate transferability, we test the accuracies of different models over adversarial examples generated from all $10000$ CIFAR-10 test images. We then show that we can select a nearly-optimal layer for transferability using only the source model.

\subsection{ILA Targeted at Different $L$ Values}

\begin{figure}[htb]
  \begin{minipage}[t]{0.4\textwidth}
    \centering
    \includegraphics[width=\textwidth,valign=t]{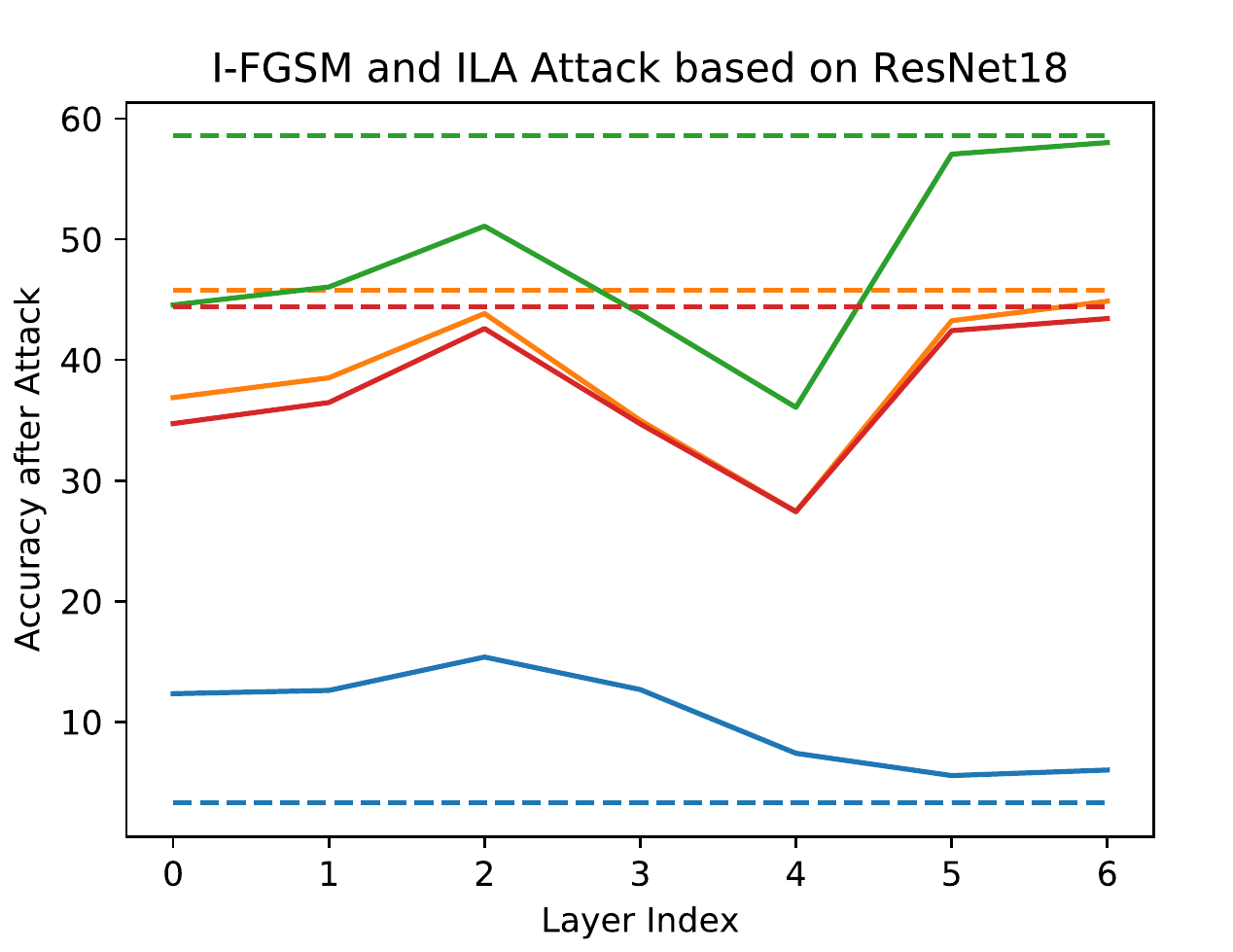}
    \caption{\label{fig:manymodeltransfer} Transfer results of ILA against I-FGSM on ResNet18 as measured by DenseNet121, SENet18, and GoogLeNet (lower accuracies indicate better attack). $\alpha = 3$.}
  \end{minipage}
  \hspace{-0.1cm}
  \begin{minipage}[t]{0.06\textwidth}
    \centering
    \vspace{0.2cm}
    \includegraphics[width=2\textwidth,valign=t]{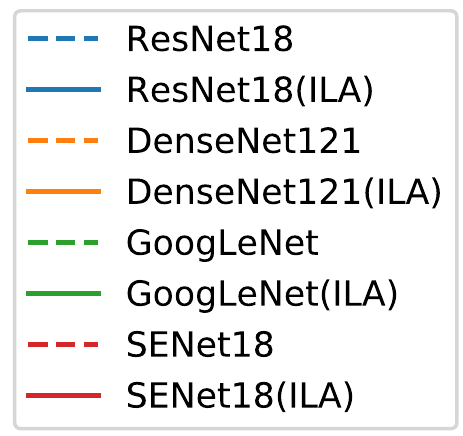}
 \end{minipage}
  \hfill
  \begin{minipage}[t]{0.45\textwidth}
    \centering
    \includegraphics[width=\textwidth,valign=t]{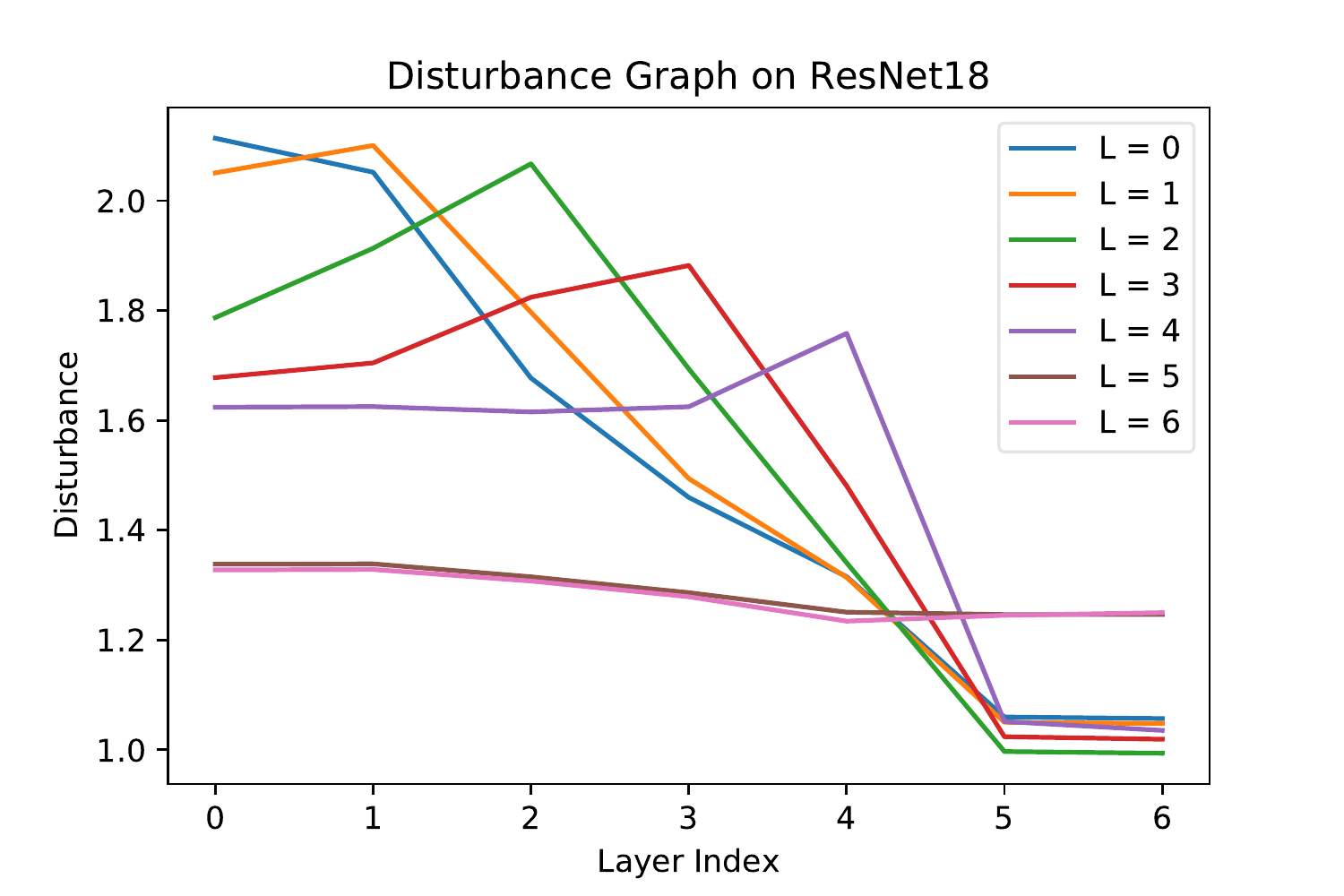}
    \caption{\label{fig:peaks} Disturbance values $\left(\frac{||\Delta Y''_l||_2}{||\Delta Y'_l||_2}\right)$ at each layer for ILA targeted at layer $l = 0,1,...,6$ for ResNet18. Observe that the $l$ in the legend refers to the hyperparameter set in the ILA attack, and afterwards the disturbance values were computed on layers indicated by the $l$ in the x-axis. Note that the last peak is produced by the $l = 4$ ILA attack.}
  \end{minipage}
\end{figure}

To confirm the effectiveness of our attack, we fix a single source model and baseline attack method, and then check how ILA transfers to the other models compared to the baseline attack. Results for ResNet18 as the source model and I-FGSM as the baseline method are shown in Figure \ref{fig:manymodeltransfer}. Comparing the results of both methods on the source model and other models, we see that ILA outperforms I-FGSM when targeting any intermediate layers, especially for the optimal hyperparameter value of $l=4$. Note that after fine-tuning, the adversarial examples perform worse on the source model, which is anticipated as ILA does not optimize for source model attack. Full results are shown in Appendix B.

\subsection{ILA with Pre-Determined $L$ Value}

Above we demonstrated that adversarial examples exhibit strongest transferability when targeting a specific layer. We wish to pre-determine this optimal value based on the source model alone so as to avoid tuning the hyperparameter $l$. To do this, we examine the relationship between transferability and the ILA layer disturbance values for a given ILA attack. We define the disturbance values of an ILA attack perturbation $X''$ as values of the function $f(l) = \frac{||\Delta Y''_l||_2}{||\Delta Y'_l||_2}$ for all values of $l$ in the source model. For each value of $l$ in ResNet18 (the set of $l$ is defined for each architecture in Appendix B) we plot the disturbance values of the corresponding ILA attack in Figure \ref{fig:peaks}. The same graph is given for other models in Appendix C. We observe that for a given source model and given ILA attack (for a specific value of $l$), the disturbance value usually reaches its peak at the targeted layer. Furthermore, we notice that the adversarial examples that produce the latest peak in the graph are typically the ones that have highest transferability for all transferred models (Table \ref{tab:main-result}). Given this observation, we propose that the latest $l$ that still exhibits a peak is a nearly optimal value of $l$ (in terms of maximizing transferability). For example, according to Figure \ref{fig:peaks}, we would choose $l=4$. Table \ref{tab:main-result} supports our claim and shows that selecting this layer gives an optimal or near-optimal attack.

Intuitively, we choose a layer with a peak because we want to choose a layer we can significantly perturb. We choose the latest layer with a peak because since the noise ($\Delta Y''_l$) we are adding at the layer is fairly unstructured (roughly random noise), passing it through the model will only dampen it. Thus, choosing the latest layer will minimize dampening and maximize the chance that the perturbation influences the model output.

\renewcommand*{\thefootnote}{\fnsymbol{footnote}}
\setlength{\tabcolsep}{4pt}
\begin{table}[!htb]
  \caption{\label{tab:main-result} ILA Accuracies (with pre-determined ILA layers)}
  \centering
  
  \begin{tabular}{cccccccc}
    \toprule
    & & \multicolumn{3}{c}{I-FGSM} & \multicolumn{3}{c}{I-FGSM Momentum} \\
    \cmidrule(r){3-5} \cmidrule(r){6-8}
    Source & Transfer & 20 Itr & 10 Itr ILA & Opt ILA  & 20 Itr & 10 Itr ILA &  Opt ILA \\
    \midrule
    & ResNet18\footnote[2]{Model that is exactly the same model as the source model.} & \textbf{3.3\%} & 7.4\% & 5.5\% (5)& \textbf{5.7\%} & 11.3\% & 7.4\% (5)\\
    ResNet18 & SENet18 & 44.4\% & \textbf{27.4\%} & \textbf{27.4\%} (4) & 33.8\% & \textbf{30.1\%} & \textbf{30.1\%} (4)\\
    ($l=4$) & DenseNet121 & 45.8\% & \textbf{27.4\%} & \textbf{27.4\%} (4) & 35.1\% & \textbf{30.3\%} & \textbf{30.3\%} (4)\\
    & GoogLeNet & 58.6\% & \textbf{36.1\%} & \textbf{36.1\%} (4)& 45.1\% & \textbf{37.8\%} & \textbf{37.8\%} (4)\\
    \midrule
    
    & ResNet18 & 36.8\% & \textbf{28.1\%} & \textbf{28.1\%} (4)& 31.0\% & \textbf{29.6\%} & \textbf{29.6\%} (4)\\
    
    SENet18 & $\text{SENet18}^\dagger$ & \textbf{2.4\%} & 8.7\% & 5.6\% (6)& \textbf{3.3\%} & 10.8\% & 6.6\% (6)\\
    
    ($l=4$) & DenseNet121 & 38.0\% & \textbf{28.3\%} & \textbf{28.3\%} (4) & 31.6\% & \textbf{29.6\%}& \textbf{29.6\%} (4)\\
    & GoogLeNet & 48.4\% & \textbf{36.3\%} & \textbf{36.3\%} (4)& 41.1\% & \textbf{37.1\%} &\textbf{37.1\%} (4)\\
    \midrule
    & ResNet18 & 45.1\% & 29.1\% & \textbf{27.7\%}(6) & 34.4\% & 30.4\% & \textbf{30.2\%}(6)  \\
    DenseNet121 & SENet18 & 43.4\% & 28.6\% & \textbf{27.8\%}(6) & 33.5\% & \textbf{29.8\%} & \textbf{29.8\%} (7)\\
    ($l=7$) & $\text{DenseNet121}^\dagger$ & 2.8\% & \textbf{2.6\%} & \textbf{2.6\%}(7) & 6.4\% & \textbf{4.3\%} & \textbf{4.3\%}(7) \\
    & GoogLeNet & 47.3\% & 31.8\% & \textbf{30.6\%}(6) & 36.3\% & \textbf{32.3\%} & \textbf{32.3\%} (7)\\
    \midrule
    & ResNet18 & 55.9\%& 35.2\%& \textbf{34.3\%}(3) & 44.6\% & 35.3\% &\textbf{34.5\%}(3) \\
    GoogLeNet & SENet18 & 53.6\% &34.5\%& \textbf{33.3\%} (3)& 43.0\% & 34.7\% & \textbf{34.2\%}(3)\\
    ($l=9$) & DenseNet121 & 48.9\% & \textbf{29.7\%} & \textbf{29.7\%}(3)& 38.9\% & \textbf{30.2\%} & \textbf{30.2\%}(9)\\
    & $\text{GoogLeNet}^\dagger$ & \textbf{0.9\%} & 1.3\% & 1.3\% (9)& \textbf{1.5\%} & 2.2\% & 2.2\% (9)\\
    \bottomrule
  \end{tabular}
\caption*{\textit{Table 1.} Accuracies after attack are shown for the models (lower accuracies indicate better attack). The hyperparameter $l$ in the ILA attack is being fixed for each source model as decided by the layer disturbance graphs (e.g. setting $l=4$ for ResNet18 since it was the last peak in Figure \ref{fig:peaks}). For I-FGSM $\alpha = 3$ and for I-FGSM with momentum $\alpha = 5$. ``Opt ILA'' refers to a 10 iteration ILA that chooses the optimal layer (determined by evaluating on transfer models). Note that on the original model, ILA does not usually beat out the baseline attack. This is expected since we are optimizing for greatest transfer, and the ILA objective does not attempt to further attack the source model.}
\end{table}

\subsection{ILA on Channels}
Further providing evidence for the effectiveness of our attack, we show that a modification of our method to attack each individual channel of a layer outputs a perturbation that exhibits the greatest transferability when attacking the most important channels (as measured by standard deviation of the channel activation values across the dataset, motivated by \cite{Polyak2015ChannelLevelAO}).

\begin{figure}[!htb]
  \centering
  \includegraphics[scale=0.4]{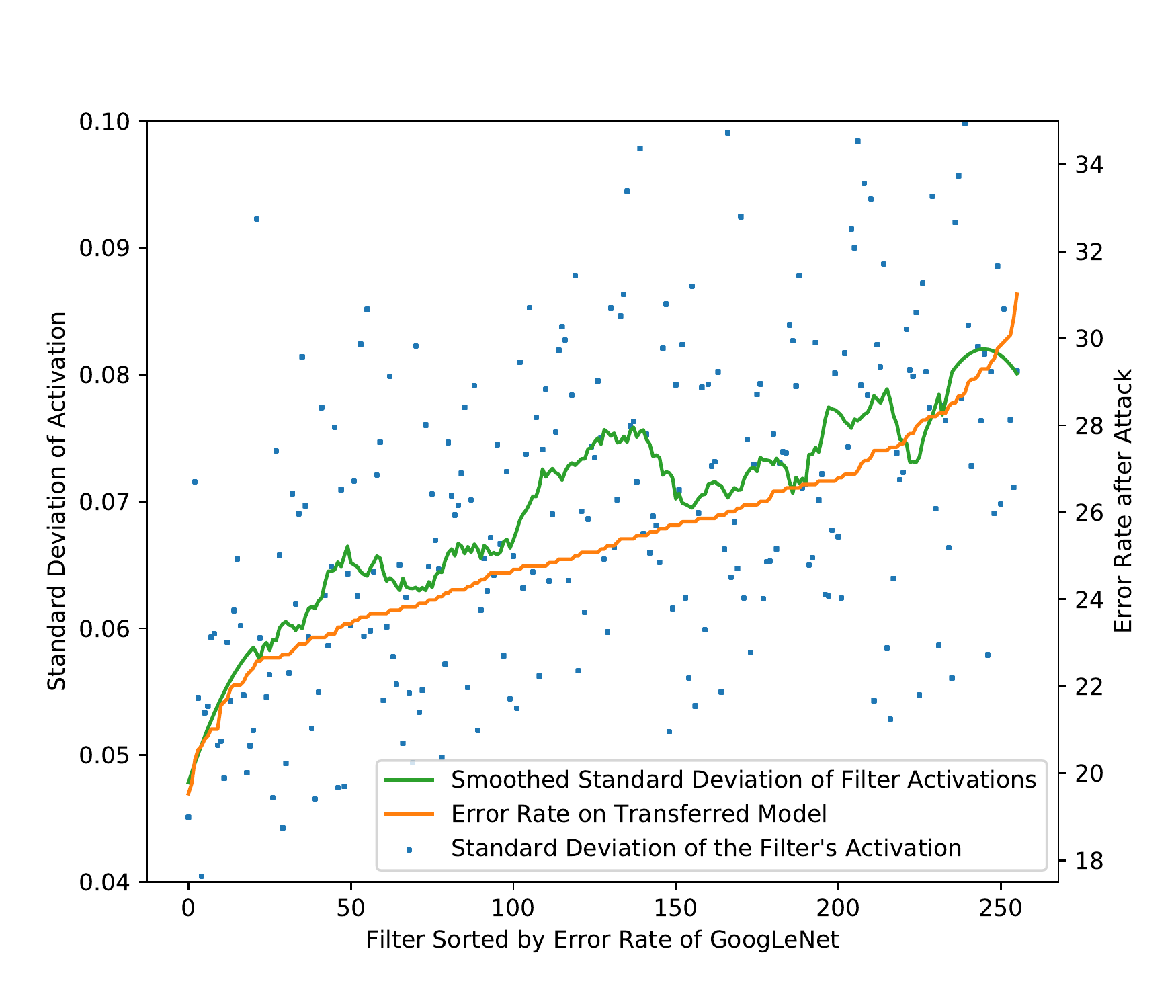}
  \caption{\label{fig:flaaux1} 
  Adversarial examples are generated on ResNet18 by targeting each of 256 channels in layer 3 of the model. We then evaluate their transferability on a GoogLeNet. In this graph, channels are sorted in order of increasing error rate on GoogLeNet. The standard deviation of activations is shown, along with the smoothed connected version (smoothed by the Savitzky-Golay filter (sliding window size 41, degree 2)).}
\end{figure}

We modify the loss function to target specific channels instead of specific layers. We do this by replacing $F_l$ in ILA as the output of a particular channel, instead of the output of a layer. The adversarial examples are generated on ResNet18 by targeting each of 256 channels in layer 3 of the model. We then evaluate their transferability on a GoogLeNet model. The result is shown in Figure \ref{fig:flaaux1}. 

The result shows that adversarial examples produced are more transferable when generated by targeting channels with higher standard deviation measured across the entire dataset. Since the variance of a channel's activations measure the information it contributes \cite{Polyak2015ChannelLevelAO}, channels with higher variance are more likely to contain information shared across models. This then shows that the increased transferability attained by attacking these channels is a result of ILA disturbing the shared information among models contributed by the channel.

\section{Related Work}

Most prior work in generating adversarial examples for attack focuses on disturbing the softmax output space via the input space \cite{Goodfellow2014ExplainingAH, Madry2017TowardsDL, MoosaviDezfooli2016DeepFoolAS, Dong2017BoostingAA}. Not many papers have focused on perturbing mid-layer outputs, though a work that has a similar goal of disturbing mid-layer activations is \cite{Mopuri2018GeneralizableDO}. It focuses on crafting a single universal perturbation that produces as many spurious mid-layer activations as possible. In contrast to our work, it does not focus on generally fine-tuning an arbitrary perturbation to become more transferable, nor does it focus on exploiting properties of a particular layer. Another work that looks at perturbing mid-layer outputs is \cite{Sabour2015AdversarialMO}. It focuses on showing that given a guide image and a very different target image it is possible to modify the target image to have a very similar embedding to that of the guide image. Unlike our work, it does not focus at all on enhancing adversarial transferability nor specific layer properties, but rather general concerns about internal deep network representations.

\section{Conclusion}
We introduce a novel attack, coined ILA, that aims to enhance the transferability of any given adversarial example. Moreover, we show that there are specific intermediate layers that we can target with ILA to substantially increase transferability with respect to regular attack baselines. We show that a near-optimal (in terms of transfer) target layer can be selected without any knowledge of specific transfer models.

Potential future work can focus on perturbing different sets of model components (i.e. different layers and channels) or further explaining the mechanism allowing ILA to improve transferability. Focusing on fine-tuning perturbations produced in different settings (i.e. generative perturbations or universal perturbations) and extending our method for targeted attack are also promising directions.


\newpage
\bibliography{fla}

\begin{thebibliography}{22}
\providecommand{\natexlab}[1]{#1}
\providecommand{\url}[1]{\texttt{#1}}
\expandafter\ifx\csname urlstyle\endcsname\relax
  \providecommand{\doi}[1]{doi: #1}\else
  \providecommand{\doi}{doi: \begingroup \urlstyle{rm}\Url}\fi

\bibitem[Athalye et~al.(2018)Athalye, Carlini, and
  Wagner]{Athalye2018ObfuscatedGG}
A.~Athalye, N.~Carlini, and D.~A. Wagner.
\newblock Obfuscated gradients give a false sense of security: Circumventing
  defenses to adversarial examples.
\newblock In \emph{ICML}, 2018.

\bibitem[Carlini and Wagner(2017)]{Carlini2017TowardsET}
N.~Carlini and D.~A. Wagner.
\newblock Towards evaluating the robustness of neural networks.
\newblock \emph{2017 IEEE Symposium on Security and Privacy (SP)}, pages
  39--57, 2017.

\bibitem[Dong et~al.(2017)Dong, Liao, Pang, Su, Zhu, Hu, and
  Li]{Dong2017BoostingAA}
Y.~Dong, F.~Liao, T.~Pang, H.~Su, J.~Zhu, X.~Hu, and J.~Li.
\newblock Boosting adversarial attacks with momentum.
\newblock 2017.

\bibitem[Eykholt et~al.(2017)Eykholt, Evtimov, Fernandes, Li, Rahmati, Xiao,
  Prakash, Kohno, and Song]{Eykholt2017RobustPA}
K.~Eykholt, I.~Evtimov, E.~Fernandes, B.~Li, A.~Rahmati, C.~Xiao, A.~Prakash,
  T.~Kohno, and D.~Song.
\newblock Robust physical-world attacks on deep learning models.
\newblock 2017.

\bibitem[Goodfellow et~al.(2014)Goodfellow, Shlens, and
  Szegedy]{Goodfellow2014ExplainingAH}
I.~J. Goodfellow, J.~Shlens, and C.~Szegedy.
\newblock Explaining and harnessing adversarial examples.
\newblock \emph{CoRR}, abs/1412.6572, 2014.

\bibitem[He et~al.(2016)He, Zhang, Ren, and Sun]{He2016DeepRL}
K.~He, X.~Zhang, S.~Ren, and J.~Sun.
\newblock Deep residual learning for image recognition.
\newblock \emph{2016 IEEE Conference on Computer Vision and Pattern Recognition
  (CVPR)}, pages 770--778, 2016.

\bibitem[Hu et~al.(2017)Hu, Shen, and Sun]{Hu2017SqueezeandExcitationN}
J.~Hu, L.~Shen, and G.~Sun.
\newblock Squeeze-and-excitation networks.
\newblock \emph{CoRR}, abs/1709.01507, 2017.

\bibitem[Huang et~al.(2017)Huang, Liu, van~der Maaten, and
  Weinberger]{Huang2017DenselyCC}
G.~Huang, Z.~Liu, L.~van~der Maaten, and K.~Q. Weinberger.
\newblock Densely connected convolutional networks.
\newblock \emph{2017 IEEE Conference on Computer Vision and Pattern Recognition
  (CVPR)}, pages 2261--2269, 2017.

\bibitem[Krizhevsky(2009)]{Krizhevsky2009LearningML}
A.~Krizhevsky.
\newblock Learning multiple layers of features from tiny images.
\newblock 2009.

\bibitem[Krizhevsky et~al.(2012)Krizhevsky, Sutskever, and
  Hinton]{Krizhevsky2012ImageNetCW}
A.~Krizhevsky, I.~Sutskever, and G.~E. Hinton.
\newblock Imagenet classification with deep convolutional neural networks.
\newblock In \emph{NIPS}, 2012.

\bibitem[Kurakin et~al.(2016)Kurakin, Goodfellow, and
  Bengio]{Kurakin2016AdversarialEI}
A.~Kurakin, I.~J. Goodfellow, and S.~Bengio.
\newblock Adversarial examples in the physical world.
\newblock \emph{CoRR}, abs/1607.02533, 2016.

\bibitem[Liu(2018)]{Liu2018}
K.~Liu.
\newblock Pytorch cifar10.
\newblock \url{https://github.com/kuangliu/pytorch-cifar}, 2018.

\bibitem[Madry et~al.(2017)Madry, Makelov, Schmidt, Tsipras, and
  Vladu]{Madry2017TowardsDL}
A.~Madry, A.~Makelov, L.~Schmidt, D.~Tsipras, and A.~Vladu.
\newblock Towards deep learning models resistant to adversarial attacks.
\newblock \emph{CoRR}, abs/1706.06083, 2017.

\bibitem[Moosavi-Dezfooli et~al.(2016)Moosavi-Dezfooli, Fawzi, and
  Frossard]{MoosaviDezfooli2016DeepFoolAS}
S.-M. Moosavi-Dezfooli, A.~Fawzi, and P.~Frossard.
\newblock Deepfool: A simple and accurate method to fool deep neural networks.
\newblock \emph{2016 IEEE Conference on Computer Vision and Pattern Recognition
  (CVPR)}, pages 2574--2582, 2016.

\bibitem[Mopuri et~al.(2018)Mopuri, Ganeshan, and
  Babu]{Mopuri2018GeneralizableDO}
K.~R. Mopuri, A.~Ganeshan, and R.~V. Babu.
\newblock Generalizable data-free objective for crafting universal adversarial
  perturbations.
\newblock \emph{IEEE transactions on pattern analysis and machine
  intelligence}, 2018.

\bibitem[Papernot et~al.(2017)Papernot, McDaniel, Goodfellow, Jha, Celik, and
  Swami]{Papernot2017PracticalBA}
N.~Papernot, P.~D. McDaniel, I.~J. Goodfellow, S.~Jha, Z.~B. Celik, and
  A.~Swami.
\newblock Practical black-box attacks against machine learning.
\newblock In \emph{AsiaCCS}, 2017.

\bibitem[Polyak and Wolf(2015)]{Polyak2015ChannelLevelAO}
A.~Polyak and L.~Wolf.
\newblock Channel-level acceleration of deep face representations.
\newblock \emph{IEEE Access}, 3:\penalty0 2163--2175, 2015.

\bibitem[Sabour et~al.(2015)Sabour, Cao, Faghri, and
  Fleet]{Sabour2015AdversarialMO}
S.~Sabour, Y.~Cao, F.~Faghri, and D.~J. Fleet.
\newblock Adversarial manipulation of deep representations.
\newblock \emph{CoRR}, abs/1511.05122, 2015.

\bibitem[Szegedy et~al.(2013)Szegedy, Zaremba, Sutskever, Bruna, Erhan,
  Goodfellow, and Fergus]{Szegedy2013IntriguingPO}
C.~Szegedy, W.~Zaremba, I.~Sutskever, J.~Bruna, D.~Erhan, I.~J. Goodfellow, and
  R.~Fergus.
\newblock Intriguing properties of neural networks.
\newblock \emph{CoRR}, abs/1312.6199, 2013.

\bibitem[Szegedy et~al.(2015)Szegedy, Liu, Jia, Sermanet, Reed, Anguelov,
  Erhan, Vanhoucke, and Rabinovich]{Szegedy2015GoingDW}
C.~Szegedy, W.~Liu, Y.~Jia, P.~Sermanet, S.~E. Reed, D.~Anguelov, D.~Erhan,
  V.~Vanhoucke, and A.~Rabinovich.
\newblock Going deeper with convolutions.
\newblock \emph{2015 IEEE Conference on Computer Vision and Pattern Recognition
  (CVPR)}, pages 1--9, 2015.

\bibitem[Tram{\`e}r et~al.(2017)Tram{\`e}r, Papernot, Goodfellow, Boneh, and
  McDaniel]{Tramr2017TheSO}
F.~Tram{\`e}r, N.~Papernot, I.~J. Goodfellow, D.~Boneh, and P.~D. McDaniel.
\newblock The space of transferable adversarial examples.
\newblock \emph{CoRR}, abs/1704.03453, 2017.

\bibitem[Zeiler and Fergus(2014)]{Zeiler2014VisualizingAU}
M.~D. Zeiler and R.~Fergus.
\newblock Visualizing and understanding convolutional networks.
\newblock In \emph{ECCV}, 2014.

\end{thebibliography}

\newpage
\renewcommand\appendixpagename{Appendix}
\appendixpage

\section*{A. Intermediate Level Attack (ILA) Algorithm}

\begin{algorithm}
\caption{Intermediate Level Attack algorithm}\label{alg:ILA}
\begin{algorithmic}[1]
\Require Original image in dataset $X$; Adversarial example $X'$ generated for $X$ by baseline attack; Function $F_l$ that calculates intermediate layer output; $L_\infty$ bound $\epsilon$; Learning rate $lr$; Iterations $n$; Loss function parameter $\alpha$.
\State $X'' = X'$\Comment{Initialize adversarial example as $X'$}
\State $i = 0$
\While{$i < n$}
\State $\Delta Y'_l = F_l(X') - F_l(X)$                             
\State $\Delta Y''_l = F_l(X'') - F_l(X)$
\State $L = \frac{\Delta Y''_l}{||\Delta Y''_l||_2} \bullet \frac{\Delta Y'_l}{||\Delta Y'_l||_2}  +  \alpha \cdot \frac{||\Delta Y''_l||_2}{||\Delta Y'_l||_2}$
\State $X'' = X'' + lr \cdot \nabla_{X''} L $
\State $X'' = clip_{\epsilon}(X'' - X) + X$ \Comment{Clip the disturbance}
\State $X'' = clip_{\text{image range}}(X'')$ \Comment{Clip image to within the natural range}
\State $i = i + 1$
\EndWhile
\State \textbf{return} $X''$
\end{algorithmic}
\end{algorithm}

\section*{B. ILA Targeted at Different $L$ Values Full Result}

As described in the main paper, we tested ILA against I-FGSM \cite{Goodfellow2014ExplainingAH} and I-FGSM with momentum \cite{Dong2017BoostingAA}. We test on a variety of models, namely: ResNet18 \cite{He2016DeepRL}, SENet18 \cite{Hu2017SqueezeandExcitationN}, DenseNet121\cite{Huang2017DenselyCC} and GoogLeNet \cite{Szegedy2015GoingDW} trained on CIFAR-10. For each source model, each large block output in the source model and each attack $A$, we generate adversarial examples for all images in the test set using $A$ with 20 iterations as a baseline. We then generate adversarial examples using $A$ with 10 iterations as input to ILA, which will then run for 10 iterations. The learning rate is set to $0.002$ for I-FGSM, $0.002$ for I-FGSM with momentum and $1.0$ for ILA. We are in the $L_\infty$ norm setting with $\epsilon = 0.03$ for all attacks. We then evaluate transferability of baseline and ILA adversarial examples over the other models by testing their accuracies. We also compare with their performance on source model, which is labeled by $\dagger$, in a similar fashion. 

Below is the list of layers (models from \cite{Liu2018}) we picked for each source model, which is indexed starting from 0 in the experiment results: 
\begin{itemize}
    \item ResNet18: conv, bn, layer1, layer2, layer3, layer4, linear (layer1-4 are basic blocks)
    \item GoogLeNet: pre\_layers, a3, b3, maxpool, a4, b4, c4, d4, e4, a5, b5, avgpool, linear
    \item DenseNet121: conv1, dense1, trans1, dense2, trans2, dense3, trans3, dense4, bn, linear
    \item SENet18: conv1, bn1, layer1, layer2, layer3, layer4, linear (layer1-4 are pre-activation blocks)
\end{itemize}

  \begin{table}[!htb]
  \caption{ Accuracies after attack using ResNet18 as source model}
  \centering
  \begin{tabular}{cccccc}
    \toprule
    Attack & Layer Index & $\text{ResNet18}^\dagger$ & SENet18 & DenseNet121  & GoogLeNet \\
    \midrule
    &0  & 12.3\% & 34.7\% & 36.9\% & 44.5\% \\
    &1  & 12.6\% & 36.5\% & 38.5\% & 46.0\% \\
    &2  & 15.4\% & 42.6\% & 43.8\% & 51.1\% \\
    ILA & 3  & 12.7\% & 34.7\% & 35.0\% & 43.8\% \\
    &4  & 7.4\% & \textbf{27.4\%} & \textbf{27.4\%} & \textbf{36.1\%}\\
    &5  & 5.5\% & 42.4\% & 43.2\% & 57.0\% \\
    &6  & 6.0\% & 43.4\% & 44.9\% & 58.0\% \\
    \midrule
    I-FGSM && \textbf{3.3\%} & 44.4\% & 45.8\% & 58.6\% \\
    \bottomrule
  \end{tabular}
  \end{table}
  
  \begin{table}[!htb]
  \caption{ Accuracies after attack using ResNet18 as source model}
  \centering
  \begin{tabular}{cccccc}
    \toprule
    Attack & Layer Index & $\text{ResNet18}^\dagger$ & SENet18 & DenseNet121  & GoogLeNet \\
    \midrule
    &0  & 18.4\% & 38.3\% & 39.8\% & 45.7\% \\
    &1  & 18.6\% & 39.4\% & 40.9\% & 46.8\% \\
    &2  & 20.0\% & 43.3\% & 44.2\% & 50.1\% \\
    ILA & 3  & 17.5\% & 37.4\% & 37.4\% & 44.2\% \\
    &4  & 11.3\% & \textbf{30.1\%} & \textbf{30.3\%} & \textbf{37.8\%} \\
    &5  & 7.4\% & 40.1\% & 40.8\% & 53.2\% \\
    &6  & 8.1\% & 41.6\% & 42.1\% & 54.7\% \\
    \midrule
    I-FGSM && \textbf{5.7\%} & 33.8\% & 35.1\% & 45.1\% \\
    Momentum & \\
    \bottomrule
  \end{tabular}
  \end{table}
  
  \begin{table}[!htb]
  \caption{ Accuracies after attack using SENet18 as source model}
  \centering
  \begin{tabular}{cccccc}
    \toprule
    Attack & Layer Index & ResNet18 & $\text{SENet18}^\dagger$ & DenseNet121  & GoogLeNet \\
    \midrule
    &0  & 33.4\% & 9.5\% & 34.7\% & 41.4\% \\
    &1  & 46.6\% & 15.0\% & 47.7\% & 53.9\% \\
    &2  & 49.6\% & 16.8\% & 49.8\% & 55.2\% \\
    ILA & 3 & 34.3\% & 11.0\% & 34.7\% & 42.5\% \\
    &4  & \textbf{28.1\%} & 8.7\% & \textbf{28.3\%} & \textbf{36.3\%} \\
    &5  & 37.3\% & 5.9\% & 38.2\% & 47.8\% \\
    &6  & 36.5\% & 5.6\% & 37.5\% & 47.6\% \\
    \midrule
    I-FGSM && 36.8\% & \textbf{2.4\%} & 38.0\% & 48.4\% \\
    \bottomrule
  \end{tabular}
  \end{table}
  
  \begin{table}[!htb]
  \caption{ Accuracies after attack using $\text{SENet18}^\dagger$ as source model}
  \centering
  \begin{tabular}{cccccc}
    \toprule
    Attack & Layer Index & ResNet18 & SENet18 & DenseNet121  & GoogLeNet \\
    \midrule
    &0  & 34.9\% & 12.1\% & 36.0\% & 42.4\% \\
    &1  & 44.9\% & 16.6\% & 45.9\% & 52.2\% \\
    &2  & 47.6\% & 18.2\% & 48.5\% & 53.4\% \\
    ILA & 3  & 35.3\% & 13.1\% & 35.5\% & 42.2\% \\
    &4  & \textbf{29.6\%} & 10.8\% & \textbf{29.6\%} & \textbf{37.1\%} \\
    &5  & 36.2\% & 6.9\% & 37.1\% & 46.3\% \\
    &6  & 35.6\% & 6.6\% & 36.3\% & 45.9\% \\
    \midrule
    I-FGSM && 31.0\% & \textbf{3.3\%} & 31.6\% & 41.1\% \\
    Momentum & \\
    \bottomrule
  \end{tabular}
  \end{table}
  
  \begin{table}[!htb]
  \caption{ Accuracies after attack using DenseNet121 as source model}
  \centering
  \begin{tabular}{cccccc}
    \toprule
    Attack & Layer Index & ResNet18 & SENet18 & $\text{DenseNet121}^\dagger$  & GoogLeNet \\
    \midrule
    &0  & 37.6\% & 37.2\% & 12.9\% & 39.5\% \\
    &1  & 54.7\% & 53.4\% & 24.6\% & 55.3\% \\
    &2  & 42.3\% & 41.5\% & 16.4\% & 45.0\% \\
    ILA & 3  & 39.4\% & 38.7\% & 14.1\% & 42.1\% \\
    &4  & 37.6\% & 36.4\% & 12.9\% & 39.9\% \\
    &5  & 30.0\% & 29.7\% & 7.1\% & 32.9\% \\
    &6  & \textbf{27.7\%} & \textbf{27.8\%} & 3.1\% & \textbf{30.6\%} \\
    &7  & 29.1\% & 28.6\% & \textbf{2.6\%} & 31.8\% \\
    &8  & 41.2\% & 39.5\% & 4.1\% & 43.4\% \\
    &9  & 44.2\% & 42.9\% & 6.4\% & 46.7\% \\
    \midrule
    I-FGSM && 45.1\% & 43.4\% & 2.8\% & 47.3\% \\
    \bottomrule
  \end{tabular}
  \end{table}
  
  \begin{table}[!htb]
  \caption{Accuracies after attack using DenseNet121 as source model}
  \centering
  \begin{tabular}{cccccc}
    \toprule
    Attack & Layer Index & ResNet18 & SENet18 & $\text{DenseNet121}^\dagger$  & GoogLeNet \\
    \midrule
    &0  & 42.3\% & 41.7\% & 20.9\% & 43.9\% \\
    &1  & 53.0\% & 52.5\% & 27.4\% & 54.2\% \\
    &2  & 44.4\% & 44.2\% & 22.1\% & 46.6\% \\
    ILA & 3  & 42.1\% & 41.7\% & 19.9\% & 44.0\% \\
    &4  & 40.5\% & 39.3\% & 18.4\% & 42.3\% \\
    &5  & 33.6\% & 33.2\% & 12.0\% & 35.7\% \\
    &6  & \textbf{30.2\%} & 30.1\% & 5.4\% & 32.4\% \\
    &7  & 30.4\% & \textbf{29.8\%} & \textbf{4.3\%} & \textbf{32.3\%} \\
    &8  & 39.2\% & 38.3\% & 5.8\% & 41.3\% \\
    &9  & 41.9\% & 40.6\% & 10.0\% & 43.8\% \\
    \midrule
    I-FGSM && 34.4\% & 33.5\% & 6.4\% & 36.3\% \\
    Momentum & \\
    \bottomrule
  \end{tabular}
  \end{table}
  
  \begin{table}[!htb]
  \caption{ Accuracies after attack using GoogLeNet as source model}
  \centering
  \begin{tabular}{cccccc}
    \toprule
    Attack & Layer Index & ResNet18 & SENet18 & DenseNet121  & $\text{GoogLeNet}^\dagger$ \\
    \midrule
    &0  & 45.7\% & 44.5\% & 42.3\% & 6.3\% \\
    &1  & 56.1\% & 55.2\% & 54.3\% & 9.4\% \\
    &2  & 58.3\% & 57.0\% & 54.8\% & 12.7\% \\
    ILA & 3  & \textbf{34.3\%} & \textbf{33.3\%} & \textbf{29.7\%} & 3.8\% \\
    &4  & 44.4\% & 42.6\% & 39.9\% & 7.8\% \\
    &5  & 42.1\% & 40.4\% & 37.9\% & 7.5\% \\
    &6  & 40.1\% & 38.4\% & 35.9\% & 6.7\% \\
    &7  & 37.0\% & 35.9\% & 33.1\% & 5.4\% \\
    &8  & 34.5\% & 33.7\% & 30.3\% & 4.1\% \\
    &9  & 35.2\% & 34.5\% & \textbf{29.7\%} & 1.3\% \\
    &10  & 58.2\% & 56.8\% & 52.7\% & 2.0\% \\
    &11  & 57.9\% & 56.1\% & 51.6\% & 2.3\% \\
    &12  & 55.8\% & 53.8\% & 49.0\% & 2.5\% \\
    \midrule
    I-FGSM && 55.9\% & 53.6\% & 48.9\% & \textbf{0.9\%} \\
    \bottomrule
  \end{tabular}
  \end{table}

   \begin{table}[!htb]
  \caption{ Accuracies after attack using GoogLeNet as source model}
  \centering
  \begin{tabular}{cccccc}
    \toprule
    Attack & Layer Index & ResNet18 & SENet18 & DenseNet121  & $\text{GoogLeNet}^\dagger$ \\
    \midrule
    &0  & 45.3\% & 44.3\% & 42.1\% & 8.9\% \\
    &1  & 53.0\% & 51.9\% & 50.9\% & 11.4\% \\
    &2  & 55.7\% & 54.4\% & 52.7\% & 14.2\% \\
    ILA & 3  & \textbf{34.5\%} & \textbf{34.2\%} & 30.6\% & 5.6\% \\
    &4  & 43.6\% & 42.5\% & 39.9\% & 9.9\% \\
    &5  & 41.5\% & 40.4\% & 38.1\% & 9.2\% \\
    &6  & 39.5\% & 38.8\% & 36.2\% & 8.6\% \\
    &7  & 37.3\% & 36.3\% & 34.1\% & 7.3\% \\
    &8  & 35.2\% & 34.6\% & 31.3\% & 6.0\% \\
    &9  & 35.3\% & 34.7\% & \textbf{30.2\%} & 2.2\% \\
    &10  & 55.4\% & 54.2\% & 49.6\% & 2.9\% \\
    &11  & 55.0\% & 53.5\% & 49.0\% & 3.1\% \\
    &12  & 53.1\% & 51.4\% & 46.9\% & 3.3\% \\
    \midrule
    I-FGSM && 44.6\% & 43.0\% & 38.9\% & \textbf{1.5\%} \\
    Momentum & \\
    \bottomrule
  \end{tabular}
  \end{table}

\begin{figure}[!htb]
  \centering
  \includegraphics[width=\textwidth,valign=t]{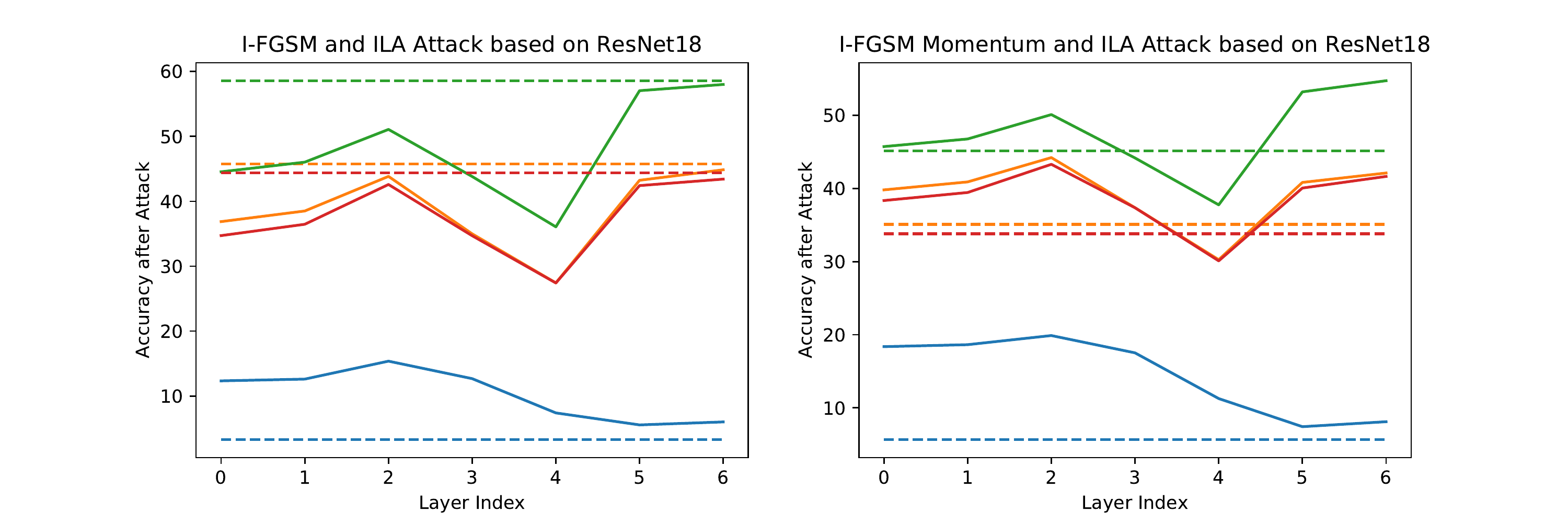}
  \includegraphics[width=\textwidth,valign=t]{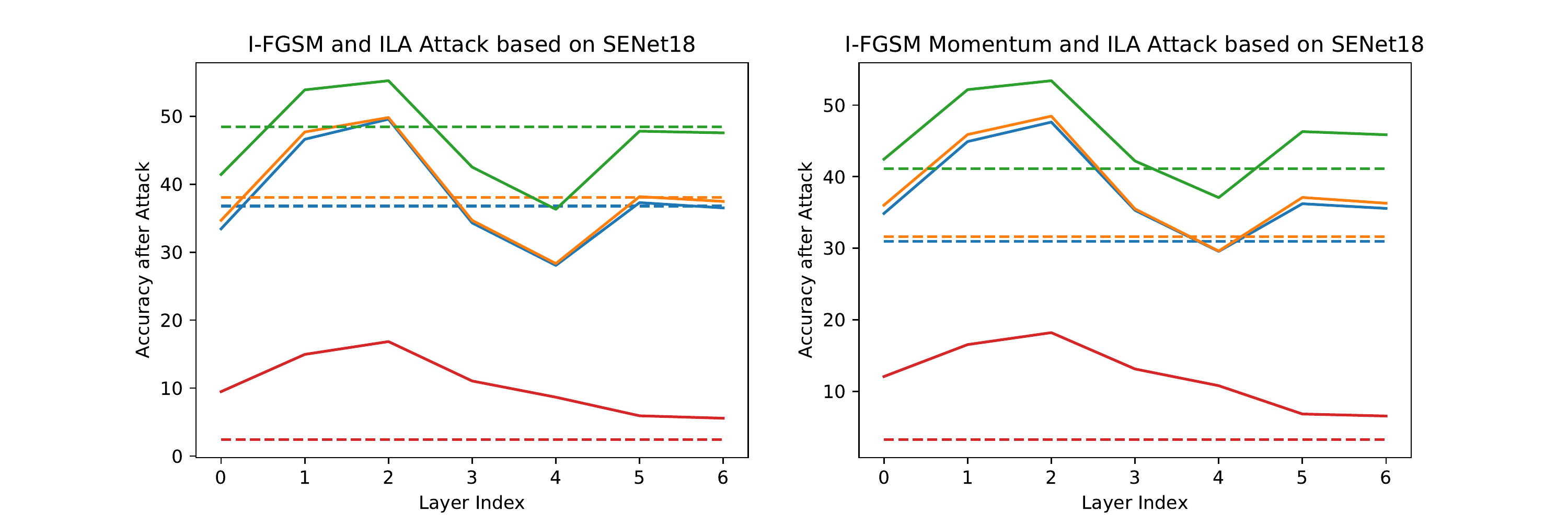}
  \includegraphics[width=\textwidth,valign=t]{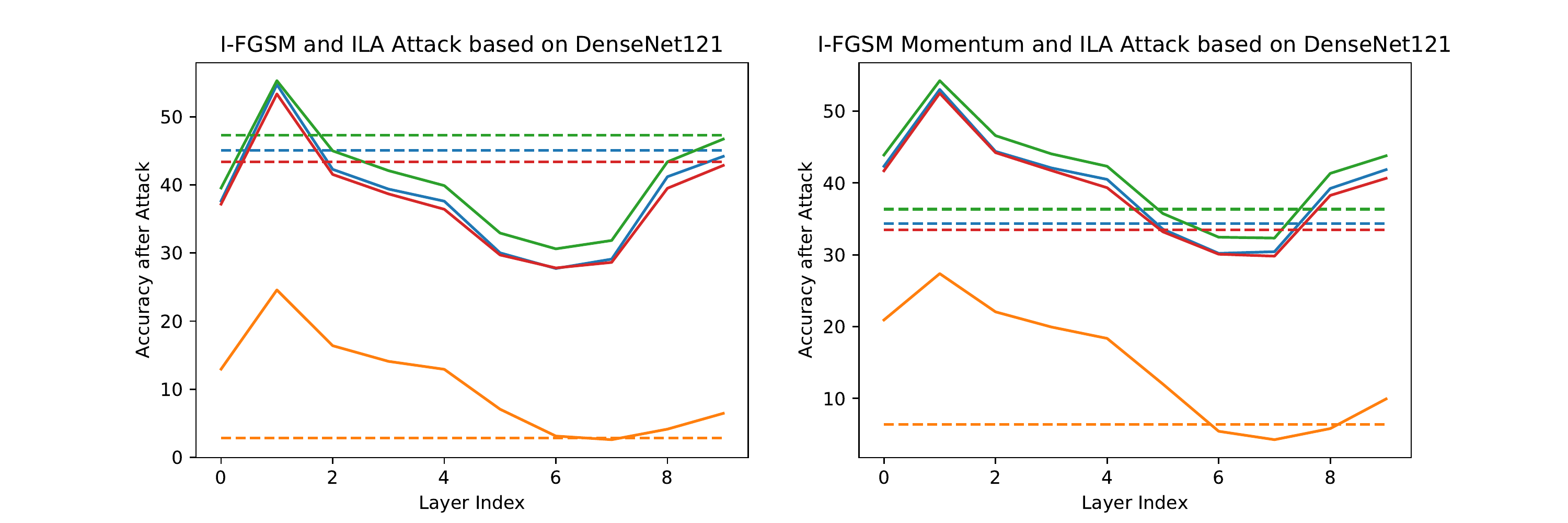}
  \includegraphics[width=\textwidth,valign=t]{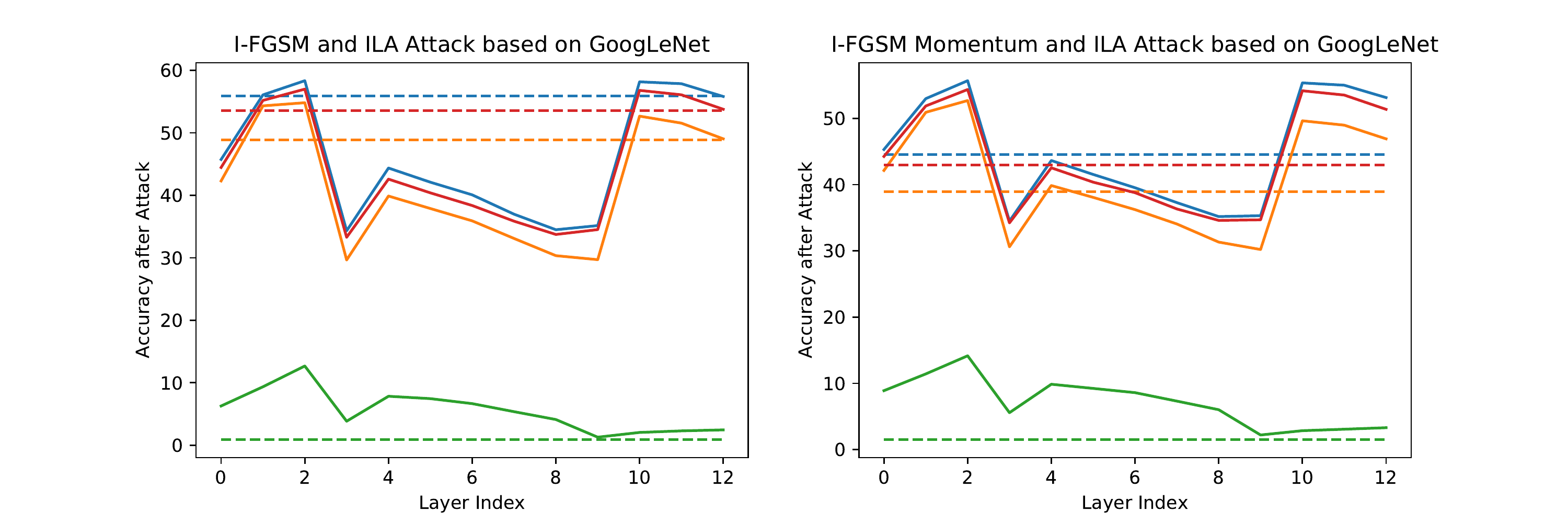}
  \includegraphics[width=0.2\textwidth,valign=t]{transferability_legend.pdf}
  \caption{\label{fig:tableVisualization} Visualizations for the data in previous tables.}
\end{figure}   

\FloatBarrier
\newpage
\section*{C. Disturbance Graphs}

In this experiment, we used the same setting as our main experiment in Appendix B to generate adversarial examples, with only I-FGSM used as the baseline attack. The average disturbance of each set of adversarial examples is calculated at each layer. We repeated the experiment for all four models described in Appendix B. Observe that the $l$ in the legend refers to the hyperparameter set in the ILA attack, and afterwards the disturbance values were computed on layers indicated by the $l$ in the x-axis.

\begin{figure}[htb]
  \begin{minipage}[t]{0.45\textwidth}
    \centering
    \includegraphics[width=\textwidth,valign=t]{ResNet18_layer_dis.pdf}
  \end{minipage}
  \hspace{-0.1cm}
  \hfill
  \begin{minipage}[t]{0.45\textwidth}
    \centering
    \includegraphics[width=\textwidth,valign=t]{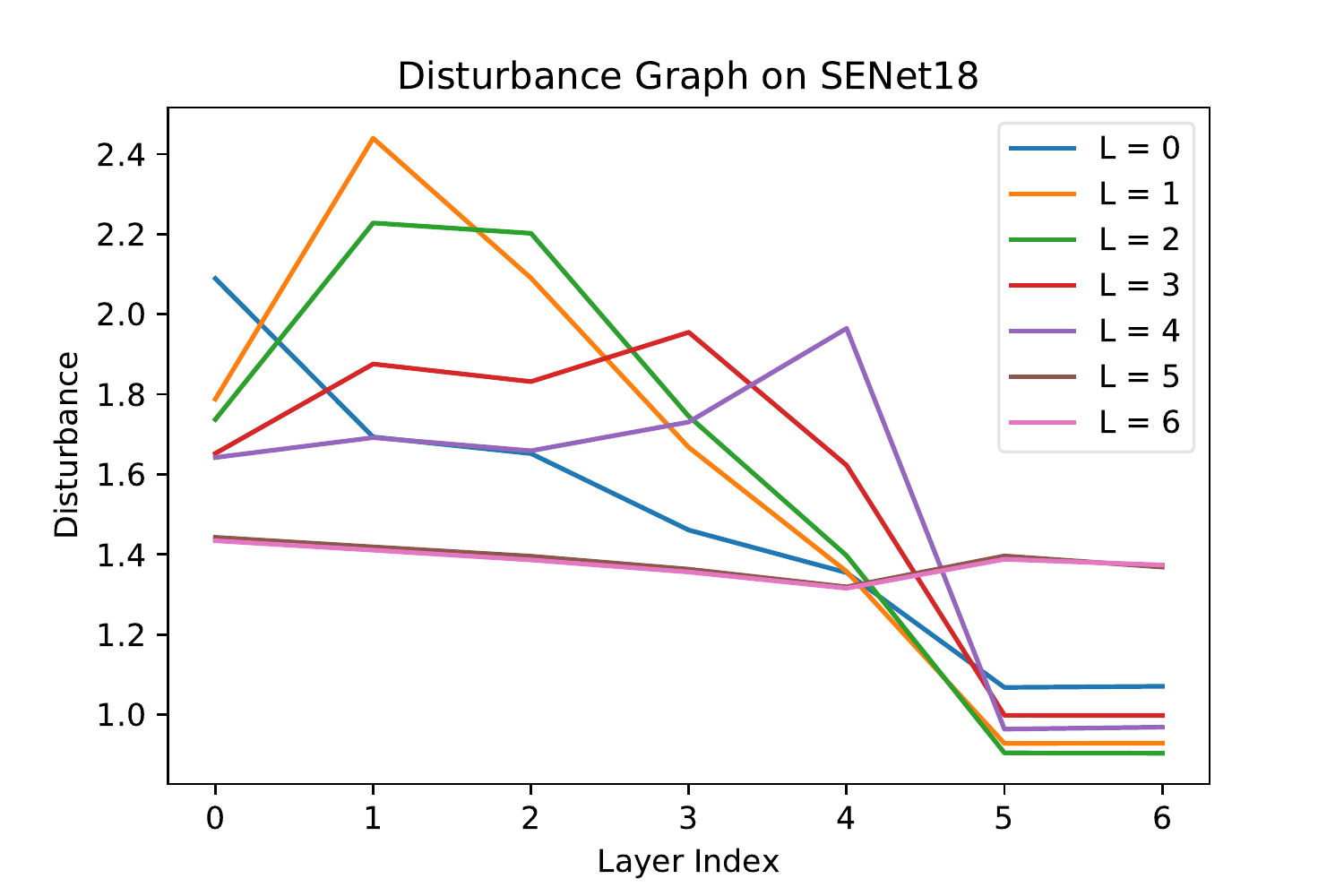}
  \end{minipage}
\end{figure}

\begin{figure}[htb]
  \begin{minipage}[t]{0.45\textwidth}
    \centering
    \includegraphics[width=\textwidth,valign=t]{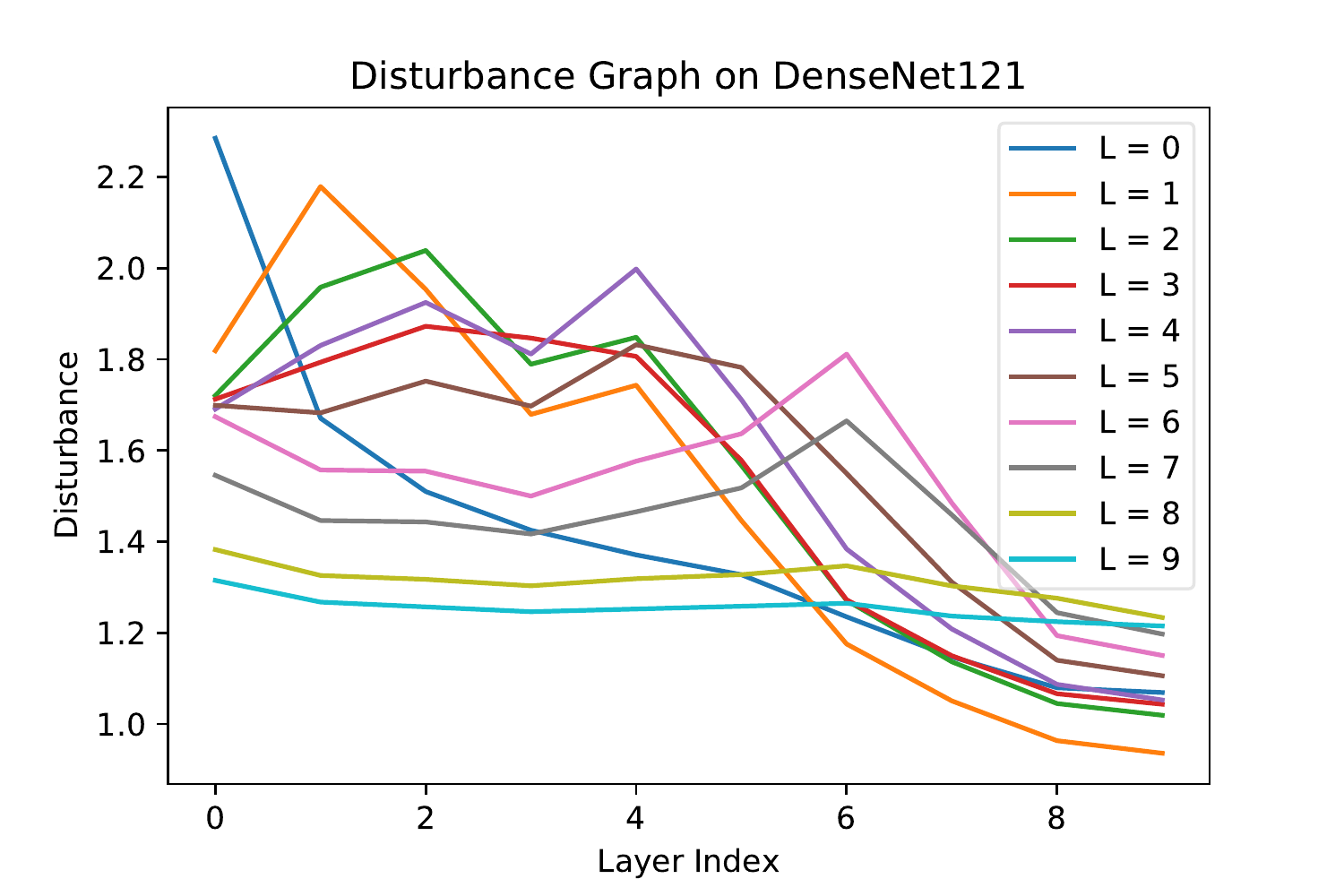}
  \end{minipage}
  \hspace{-0.1cm}
  \hfill
  \begin{minipage}[t]{0.45\textwidth}
    \centering
    \includegraphics[width=\textwidth,valign=t]{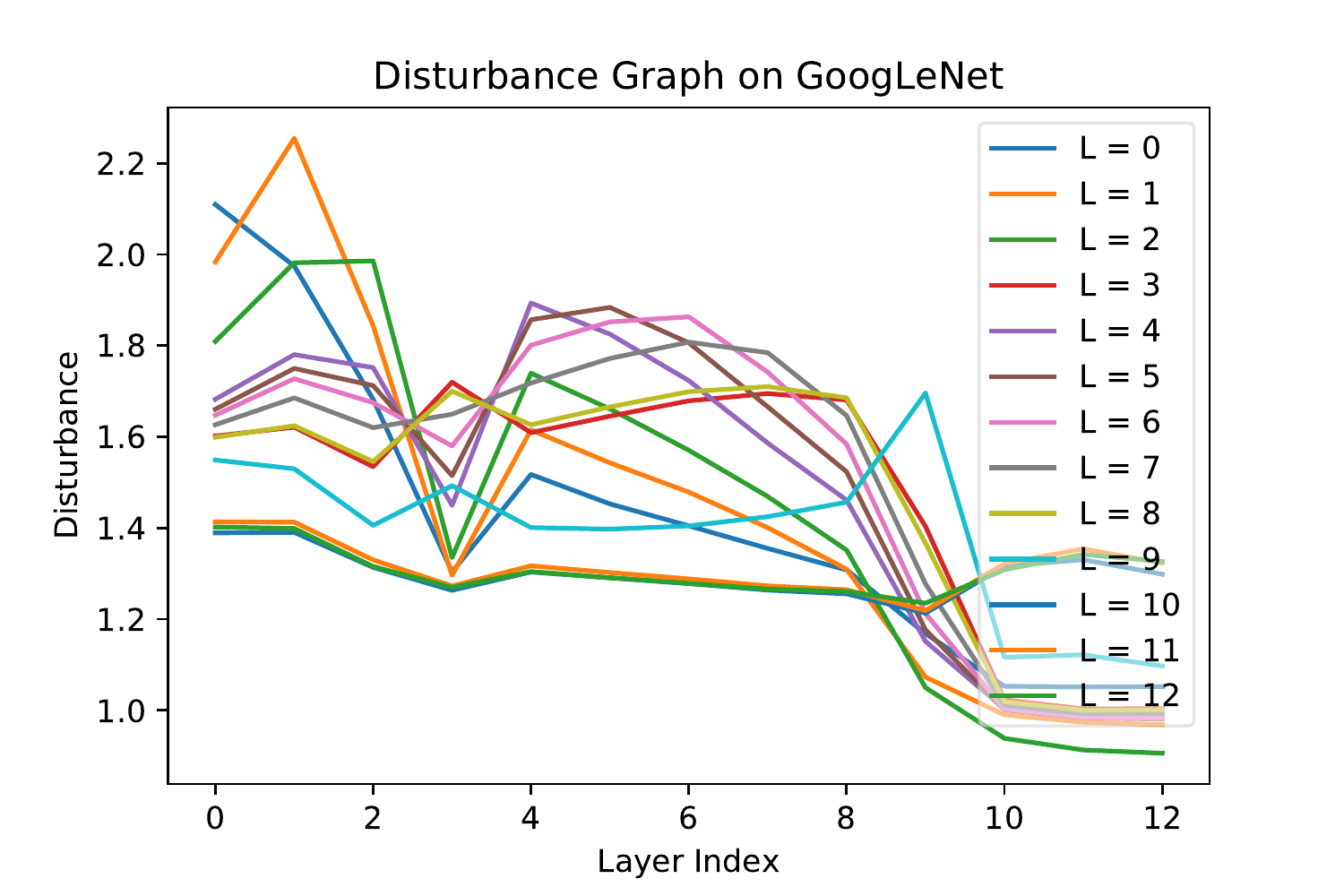}
  \end{minipage}
  \caption{\label{fig:LayerLossGraphs} Layer disturbance graphs on different source models.}
\end{figure}

\newpage
\section*{D. Fooling with Different $L_\infty$ Values}
In this experiment, we use ILA to generate adversarial examples with an I-FGSM baseline attack on ResNet18 with $\epsilon = 0.02, 0.03, 0.04, 0.05$. We then evaluated their transferability against I-FGSM baseline on the generated $10 \times 32 = 320$ adversarial examples.

\begin{figure}[htb]
  \begin{minipage}[t]{0.45\textwidth}
    \centering
    \includegraphics[width=\textwidth,valign=t]{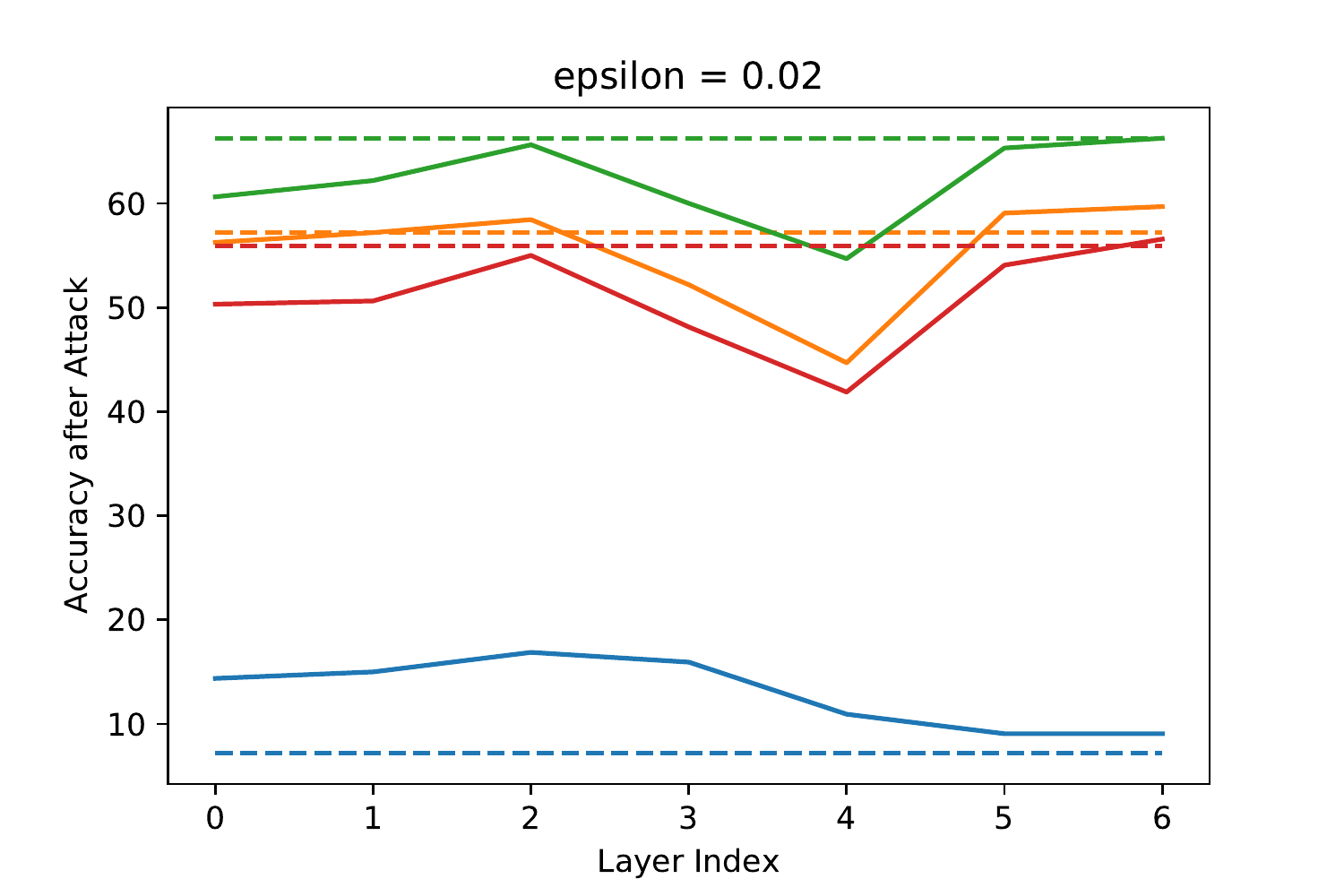}
  \end{minipage}
  \hspace{-0.1cm}
  \hfill
  \begin{minipage}[t]{0.45\textwidth}
    \centering
    \includegraphics[width=\textwidth,valign=t]{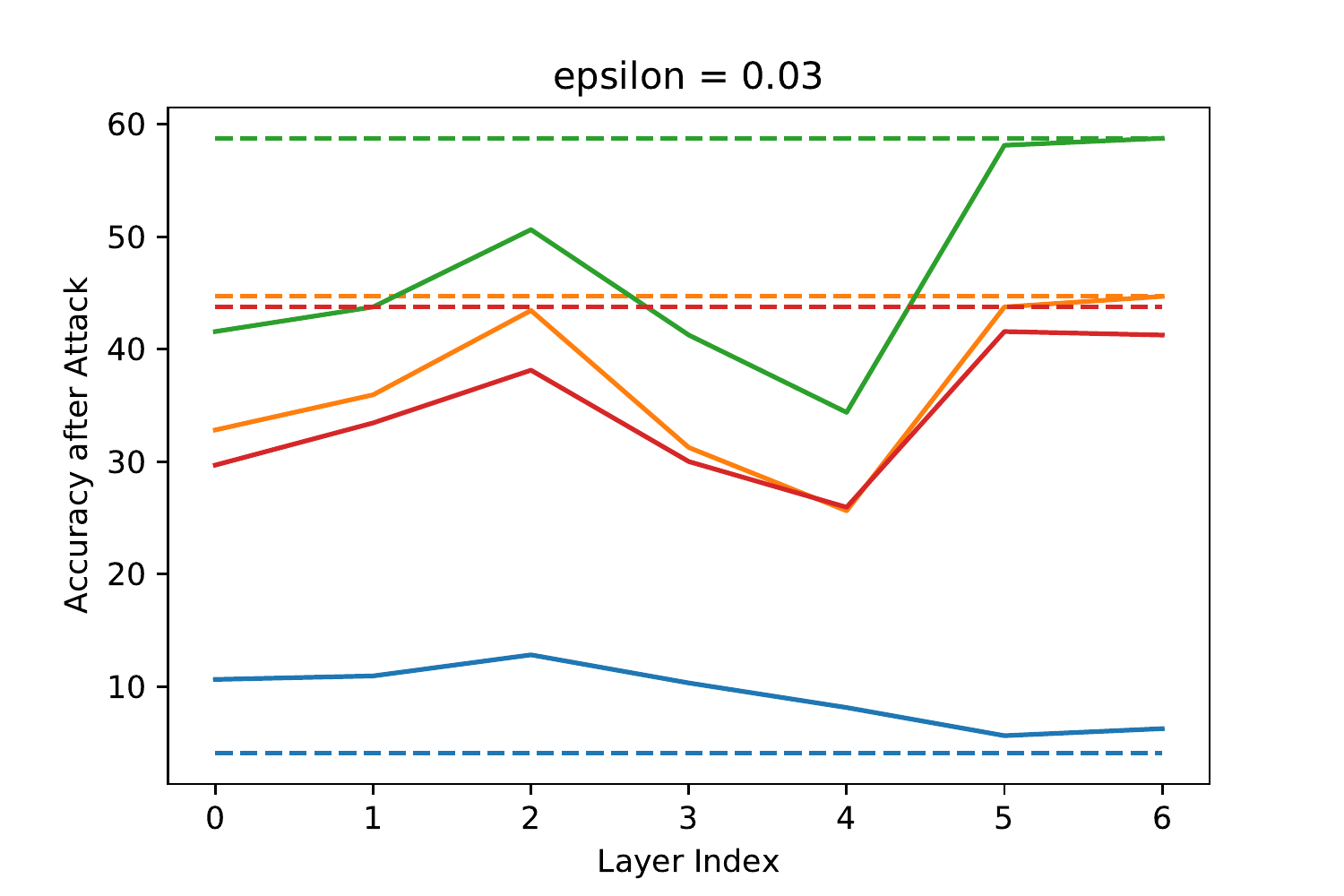}
  \end{minipage}
\end{figure}

\begin{figure}[htb]
  \begin{minipage}[t]{0.45\textwidth}
    \centering
    \includegraphics[width=\textwidth,valign=t]{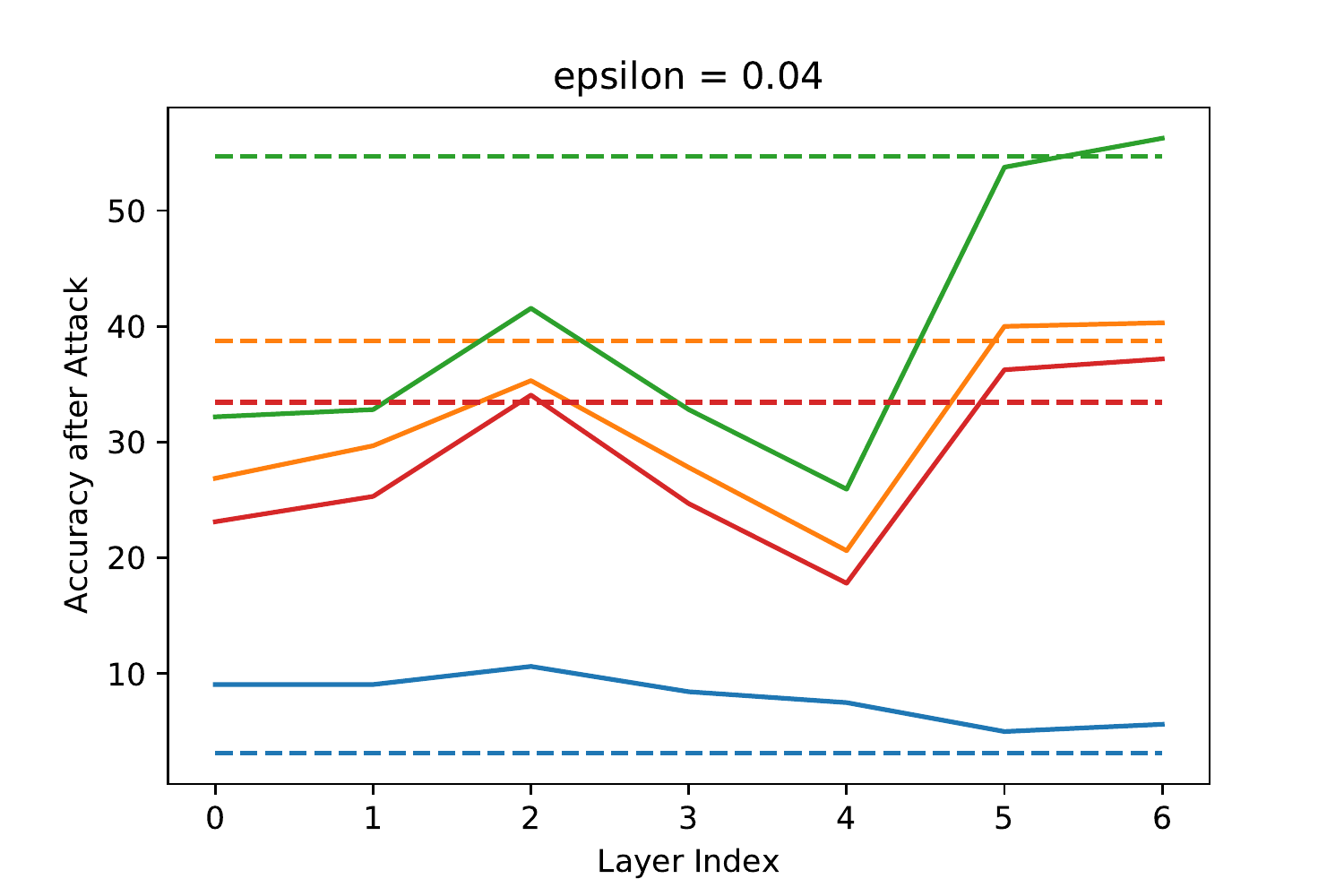}
  \end{minipage}
  \hspace{-0.1cm}
  \hfill
  \begin{minipage}[t]{0.45\textwidth}
    \centering
    \includegraphics[width=\textwidth,valign=t]{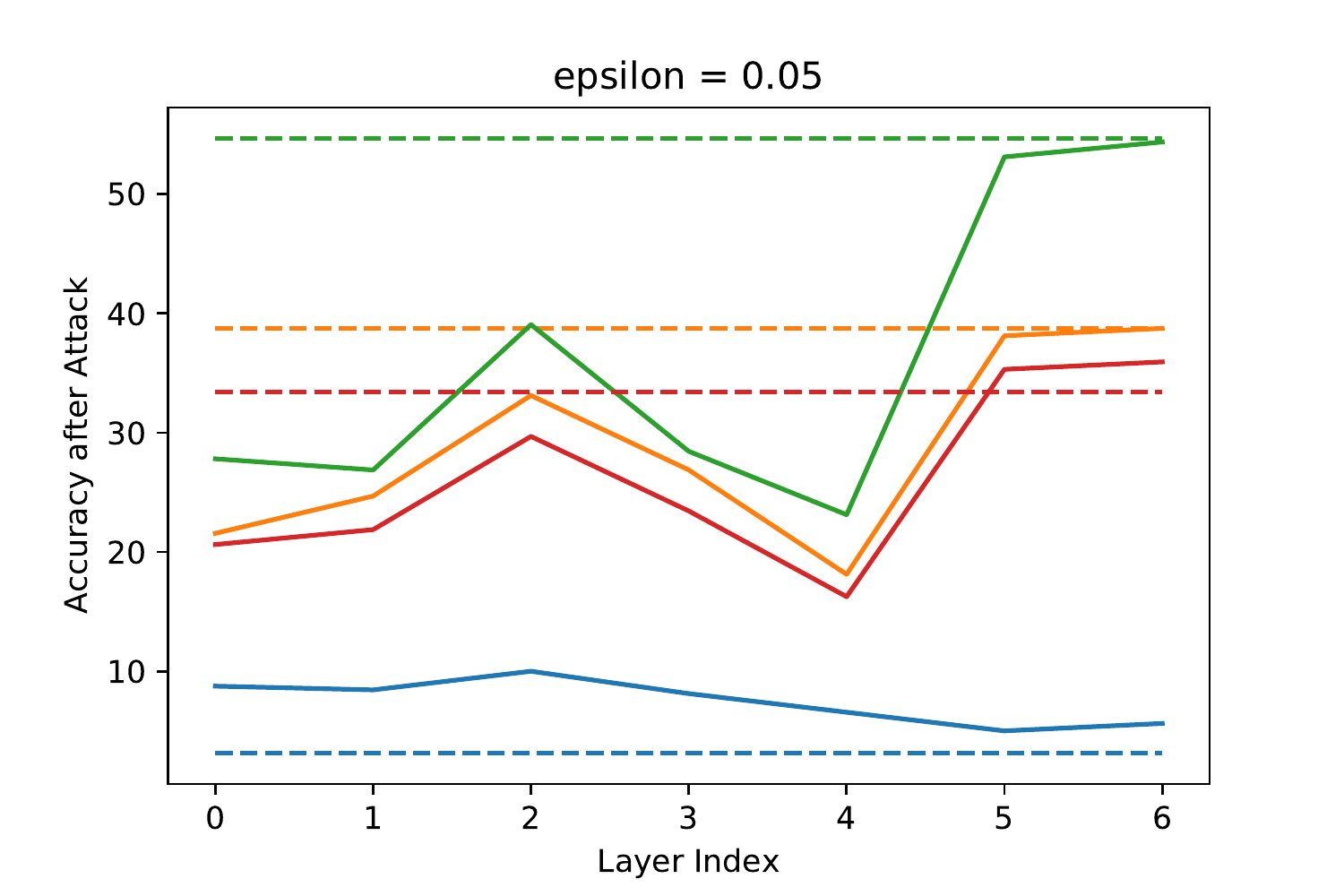}
  \end{minipage}
  \centering
   \includegraphics[width=0.2\textwidth,valign=t]{transferability_legend.pdf}
   \caption{\label{fig:epsilons} Transferability graphs for different epsilons}
\end{figure}
\hspace{2em}
\newpage
\section*{E. Learning Rate Ablation}

We set iterations to 20 for both I-FGSM and I-FGSM with Momentum and experimented different learning rates on ResNet18. We then evaluate different models' accuracies on the generated $50 \times 32 = 1600$ adversarial examples. 

\begin{table}[!th]
  \caption{\label{tab:ifgsmlr} I-FGSM }
  \centering
  \begin{tabular}{cccccc}
    \toprule
    learning rate &  ResNet18\footnote[2]{Model that is exactly the same model as the source model.} & SENet18 & DenseNet121 & GoogLeNet \\
    \midrule
    0.002  & 3.3\% & 44.9\% & 47.1\% & 59.3\% \\
    0.008  & 0.8\% & 45.6\% & 46.8\% & 60.0\% \\
    0.014  & 0.6\% & 47.2\% & 49.4\% & 59.5\% \\
    0.02 & 1.3\%  & 46.8\% & 51.4\% & 59.8\% \\
    \bottomrule
  \end{tabular}
  \end{table}
  
  \begin{table}[h]
  \caption{\label{tab:momentumifgsmlr} I-FGSM with Momentum }
  \centering
  \begin{tabular}{cccccc}
    \toprule
    learning rate &  $\text{ResNet18}^\dagger$ & SENet18 & DenseNet121 & GoogLeNet \\
    \midrule
    0.002  & 5.9\% & 35.0\% & 36.6\% & 46.1\% \\
    0.008  & 0.6\% & 43.0\% & 43.8\% & 56.1\% \\
    0.014  & 0.4\% & 43.6\% & 45.2\% & 55.9\% \\
    0.02 & 0.4\%  & 44.1\% & 46.4\% & 57.2\% \\
    \bottomrule
  \end{tabular}
  \end{table}

\end{document}